\documentclass{article}

\usepackage{PRIMEarxiv}

\usepackage[utf8]{inputenc} 
\usepackage[T1]{fontenc}    
\usepackage{hyperref}       
\usepackage{url}            
\usepackage{booktabs}       
\usepackage{amsfonts}       
\usepackage{nicefrac}       
\usepackage{microtype}      
\usepackage{lipsum}
\usepackage{fancyhdr}       
\usepackage{graphicx}       
\usepackage{amsmath}
\usepackage{graphicx}
\usepackage{url} 
\usepackage{siunitx}
\usepackage{xcolor}
\usepackage{rotating}
\usepackage{float}
\usepackage{flafter}
\usepackage{longtable}
\usepackage[title]{appendix}

\title{Tuning support vector machines and boosted trees using optimization algorithms
\thanks{\textit{\underline{Citation}}: 
\textbf{Lundell, Jill F., Tuning support vector machines and boosted trees using optimization algorithms.}} 
}

\author{
 Jill Lundell \\
  Department of Data Science\\
  Dana-Farber Cancer Institute\\
  Department of Biostatistics\\
  Harvard T.H. Chan School of Public Health\\
  Boston, MA 02215 \\
  \texttt{jlundell@ds.dfci.harvard.edu} \\
}

\begin{document}
\maketitle

\begin{abstract}
Statistical learning methods have been growing in popularity in recent years. Many of these procedures have parameters that must be tuned for models to perform well. Research has been extensive in neural networks, but not for many other learning methods. We looked at the behavior of tuning parameters for support vector machines, gradient boosting machines, and adaboost in both a classification and regression setting. We used grid search to identify ranges of tuning parameters where good models can be found across many different datasets. We then explored different optimization algorithms to select a model across the tuning parameter space. Models selected by the optimization algorithm were compared to the best models obtained through grid search to select well performing algorithms. This information was used to create an R package, \texttt{EZtune}, that automatically tunes support vector machines and boosted trees. 
\end{abstract}


\section{Introduction}
\label{sec:intro}

Statistical learning models have gained in popularity in recent years because of their ability to provide greater predictive accuracy than traditional statistical models in many situations. Some models, such as Random Forests, perform well without parameter tuning, but most learning methods have parameters that must be tuned \cite{breiman2001random}. The No Free Lunch theorems state that there is no one type of model that outperforms all other models in all situations \cite{schumacher2001no}. Thus, it is essential to explore several different types of models to find an optimal solution to a problem. Support vector machines (SVMs) \cite{cortes1995support}, gradient boosting machines (GBMs) \cite{gbmpaper}, and adaboost \cite{freund1997decision} are three supervised learning models that can be very accurate when tuned. However, parameters can be difficult to tune and recommendations for tuning methods in journal articles, blogs, and FAQs are often not well justified. Better understanding of the properties of tuning parameters and how to tune them is needed to effectively use these models. Software tools that allow users to tune models without requiring the user to do substantial research are lacking. Further development of tuning software would provide many data analysts with a wide range of more accessible tools for modeling. 

We explore tuning parameters for SVM, GBM, and adaboost to identify tuning parameter spaces that yield accurate predictive models. We then search over the tuning parameter space using different optimization algorithms to assess the ability of each algorithm to find a well tuned model over many datasets. This information is used to create an R package called \texttt{EZtune} that autotunes SVMs, GBMs, and adaboost \cite{Eztune} \cite{dissertation}. 

\section{Overview of tuning parameters}
\label{sec:over_tune}
The following overview of SVMs and boosted trees briefly summarizes SVMs, GBMs, and adaboost and identifies the tuning parameters for each model. 

\subsection{Support Vector Machines}
\label{sec:svm}
SVMs use separating hyperplanes to create decision boundaries for classification and regression models \cite{cortes1995support}. The separating hyperplane is called a soft margin because it allows some points to be on the wrong side of the hyperplane. The cost parameter, $C$, dictates the tolerance for points to be on the wrong side of the margin. A large value of $C$ allows many points to be on the wrong side while smaller values of $C$ have a much lower tolerance for misclassified points. A kernel, $K$, is used to map the classifier into a higher dimensional space. Hyperplanes are used to classify in the higher dimensional space, which results in non-linear boundaries in the original space. The SVM is modeled as: 

\begin{equation}
    f(x) = \beta_0 + \sum_{i \in S} \alpha_i K(x, x_i;\gamma)
\end{equation}

\noindent where, $K$ is a kernel with tuning parameter $\gamma$, $S$ is the set of support vectors (points on the boundary of the margin), and $\alpha_i$ computed using $C$ and the margin. The tuning parameters for SVM classification are $C$ and $\gamma$. Common kernels are polynomial, radial, and linear. We use the radial kernel for this work because it performs well in many situations. 

Support vector regression (SVR) has an additional tuning parameter, $\epsilon$. SVR attempts to find a function, or hyperplane, such that the deviations between the hyperplane and the responses, $y_i$, are less than $\epsilon$ for each observation \cite{smola2004tutorial}. The cost represents the number of points that can be further than $\epsilon$ away from the hyperplane. Essentially, SVMs maximize the number of points that are on the correct side of the margin and SVR maximizes the number of points that fall within $\epsilon$ of the margin.  The only mathematical restriction for the tuning parameters for SVM and SVR is that they are greater than 0. 

\subsection{Boosted Trees}
\label{sec:boost}

Boosted trees are part of the family of ensemble methods which combine weak classifiers into a single, accurate, classifier. A weak learner typically does not perform well alone, but combining many weak learners can create a strong classifier \cite{friedman2009elements}. With boosted trees, a small tree computed from the training data is the weak learner. The misclassified points or residuals from the tree are examined and the information is used to fit a new tree. The model is updated by adding the new tree to the previously fitted trees. The ensemble is iteratively updated in this manner and final predictions are made by a weighted vote of the weak learners. 

The primary difference between boosted tree algorithms is the method used to learn from misclassified observations at each iteration. Adaboost fits a small tree to the training data while applying the same weight to all observations in the dataset \cite{freund1997decision}. The misclassified points are then given greater weight than the correctly classified points and a new tree is computed. The new tree is added to the previous tree with weights. The process is repeated many times with misclassified points being given greater weight and a new tree is created using the weighted data and added to the previous model. Weak learners that perform better are given greater weight than those that classify more poorly. This results in an additive model where the final predictions are the weighted sum of the predictions made by all of the models in the ensemble \cite{friedman2009elements}. 

GBMs are a boosted tree that uses gradient descent to minimize a loss function during the learning process \cite{gbmpaper}. The loss function can be tailored to the problem being solved. The MSE was used as the loss function for regression models and a logarithmic loss was used for classification problems in this article. A decision tree is used as the weak learner and trees are kept small to ensure that they are weak. GBMs recursively fit new trees to the residuals from previous trees and then combine the predictions from all of the trees to obtain a final prediction.  

Adaboost and GBMs have a nearly identical set of tuning parameters. Both models require that the number of iterations, depth of the trees, and the shrinkage, which controls how fast the trees learn, are tuned. GBMs have an additional tuning parameter which is the minimum number of observations in the terminal nodes.


\section{Methods}
\label{sec:methods}

Tuning parameters were analyzed using six datasets with a binary response and seven with a continuous response. Table \ref{tbl:bindat} lists the datasets and their characteristics. An extensive grid search was done for each dataset to understand the response surface of the mean squared error (MSE) or classification accuracy across tuning parameter values. The response surface was used to to identify a tuning parameter space that contained several of the best fitting models for each of the data sets. Optimization algorithms searched through the tuning parameter spaces to find a well tuned model. The MSEs or classification accuracies of the optimization algorithms were compared to the best fitting model found in the grid search to determine which optimization algorithms performed well both in general and relative to each other. The error measure and computation time were both considered when assessing the performance of each optimization algorithm. Computation time was included as a performance measure to ensure that implementation of selected algorithms is practical.

\begin{table}[ht]
\caption{List of datasets used to explore tuning parameters.}
\begin{center}
    \begin{tabular}{lSSSS}
    \hline
    \multicolumn{1}{c}{	} &	{ } & { } 	&	{Number of}	&	{Number of}	\\
    \multicolumn{1}{c}{	} &	{ } & { } 	&	{Categorical}	&	{Continuous}	\\
    \multicolumn{1}{c}{	} &	{Number of} & {Number of} 	&	{Explanatory}	&	{Explanatory}	\\
    \multicolumn{1}{c}{	} &	{Dataset} & {Observations} 	&	{Variables}	&	{Variables}	\\
    \hline									
    \multicolumn{5}{l}{Binary}									\\
    \hline									
    Breast Cancer Data$^1$	&	699	&	10	&	0	&	9	\\
    Ionosphere$^1$	&	351	&	35	&	1	&	33	\\
    Pima Indians$^1$	&	768	&	9	&	0	&	8	\\
    Sonar$^1$	&	208	&	61	&	0	&	60	\\
    Lichen$^2$	&	840	&	40	&	2	&	31	\\
    Mullein$^2$	&	12094	&	32	&	0	&	31	\\
    \hline									
    \multicolumn{5}{l}{Continuous}									\\
    \hline									
    Abalone$^3$	&	4177	&	9	&	1	&	7	\\
    Boston Housing 2$^1$	&	506	&	19	&	1	&	15	\\
    CO2$^4$	&	84	&	5	&	3	&	1	\\
    Crime$^5$	&	47	&	14	&	1	&	12	\\
    Ames Housing$^6$	&	1460	&	61	&	36	&	24	\\
    Union$^5$	&	50	&	4	&	0	&	3	\\
    Wage$^5$	&	39	&	10	&	0	&	9	\\
    \hline									
    \multicolumn{5}{l}{1. mlbench \cite{mlbench}}									\\
    \multicolumn{5}{l}{2. EZtune \cite{Eztune}}									\\
    \multicolumn{5}{l}{3. AppliedPredictiveModeling \cite{apm}}									\\
    \multicolumn{5}{l}{4. datasets \cite{R}}									\\
    \multicolumn{5}{l}{5. Practicing Statistics \cite{kuiper2013practicing}}									\\
    \multicolumn{5}{l}{6. Kaggle \cite{de2011ames, kaggle}	}									\\
    \hline									
    \end{tabular}
\end{center}
\label{tbl:bindat}
\end{table}


\section{Tuning parameter spaces}
\label{sec:tune}

We searched blog posts, books, and journal articles to determine the tuning parameter ranges that have been used by other authors. Many authors do not address all of the tuning parameters and parameter space selection was not backed up by computation in any resources that we found. We used all of the tuning parameters identified in Section \ref{sec:over_tune} and did a grid search over a wide range of tuning parameter values. These ranges include all of the regions seen in other resources and are extended even further for some tuning parameters. Tables \ref{tbl:gridrangeB} and \ref{tbl:gridrangeC} show the ranges that were used for the grid search along with the tuning parameter spaces we recommend. The results of the grid search indicated that the grids were sufficiently large and do not need to be expanded beyond the limits shown in Tables \ref{tbl:gridrangeB} and \ref{tbl:gridrangeC}.  

\begin{table}[ht]
\caption{Tuning parameter ranges for models with a binary response.}
\begin{center}
    \begin{tabular}{lllll}
    \hline
    \multicolumn{1}{c}{	} &	{ } & {Grid} 	&	{Tuning}	&	{Search}	\\
    \multicolumn{1}{c}{	} &	{ } & {Search} 	&	{Parameter}	&	{Start}	\\
    \multicolumn{1}{c}{Model} &	{Parameter} & {Range} 	&	{Space}	&	{Location}	\\
    \hline									
    Support vector	&	Cost	&	[$2^{-10}$, $2^{25}$]	&	[1, 1024]	&	10	\\
    machines	&	$\gamma$	&	[$2^{-25}$, $2^{10}$]	&	[$2^{-10}$, $2^{10}$]	&	$2^{-5}$	\\
    \hline									
    Gradient	&	Number of trees	&	[50, 20,000]	&	 [50, 3000]	&	500	\\
boosting	&	Interaction depth	&	[1, 19]	&	[1, 15]	&	5	\\
machines	&	Shrinkage 	&	[0.001, 0.1]	&	[0.001, 0.1]	&	0.1	\\
    	&	Minimum number of 	&	[5, 15]	&	[5, 12]	&	8	\\
    	&	observations in	&		&		&		\\
    	&	terminal nodes	&		&		&		\\
    \hline									
    Adaboost	&	Number of trees	&	[100, 1400]	&	[50, 500]	&	300	\\
    	&	Interaction depth	&	[1, 20]	&	[1, 10]	&	10	\\
    	&	Shrinkage ($\nu$) 	&	[0.01, 1]	&	[0.01, 0.5]	&	0.05	\\
    \hline
    \end{tabular}
\end{center}
\label{tbl:gridrangeB}
\end{table}

\begin{table}[ht]
\caption{Tuning parameter ranges for models with a continuous response.}
\begin{center}
    \begin{tabular}{lllll}
    \hline
	&		&	Grid 	&	Tuning	&	Search	\\
	&		&	Search	&	Parameter	&	Start	\\
 Model	&	Parameter	&	Range	&	Space	&	Location	\\
 \hline									
 Support vector	&	Cost	&	[$2^{-10}$, $2^{25}$]	&	[1, 1024]	&	2	\\
 machines	&	$\gamma$	&	[$2^{-25}$, $2^{10}$]	&	[$2^{-10}$, $2^{0}$]	&	$2^{-5}$	\\
	&	$\epsilon$	&	[0, 10]	&	[0, 0.5]	&	0.4	\\
 \hline									
 Gradient	&	Number of trees	&	[50, 20,000]	&	 [50, 5000]	&	2000	\\
boosting	&	Interaction depth	&	[1, 19]	&	[1, 15]	&	8	\\
machines	&	Shrinkage 	&	[0.001, 0.1]	&	[0.001, 0.1]	&	0.1	\\
 	&	Minimum number of 	&	[5, 15]	&	[5, 10]	&	5	\\
 	&	observations in	&		&		&		\\
 	&	terminal nodes	&		&		&		\\
    \hline
    \end{tabular}

\end{center}
\label{tbl:gridrangeC}
\end{table}

SVM, GBM, and adaboost models were computed for tuning parameters throughout their specified grid region. The error measure was evaluated at each grid location using 10-fold cross-validation for most datasets and and 3-fold cross-validation for the largest datasets. The error measure is the misclassification rate for binary response variables and the MSE for continuous response variables. Computation time was recorded for each of the models to aid in finding regions that are computationally practical. We were able to identify parameter spaces for each model type that included accurate models and faster computation times for each dataset. The parameter space for each model type is identified in Section \ref{sec:tune} and specified in Tables \ref{tbl:gridrangeB} and \ref{tbl:gridrangeC}. Calculations were done using the packages \texttt{e1071} \cite{e1071}, \texttt{gbm} \cite{gbm}, and \texttt{ada} \cite{ada} in the R statistical software environment for statistical computing and graphics \cite{R}. Note that regression models were not computed with adaboost because the adaboost package that was used does not do regression. 

We wanted to ensure that selected tuning parameter spaces produced models with stable performance. Thus, error measures obtained from each fold of cross-validation were used to compute a 95\% upper confidence limit for the error measures to assess the consistency of error measures across the grid. The upper confidence limits were graphed at each grid location in addition to the cross validated error measure and the computation time. In all cases, the surfaces produced by the upper confidence limits yielded nearly identical surface patterns as the error measures so they are not discussed further. 

\begin{figure}
	\begin{center}
		\includegraphics[width=0.95\textwidth]{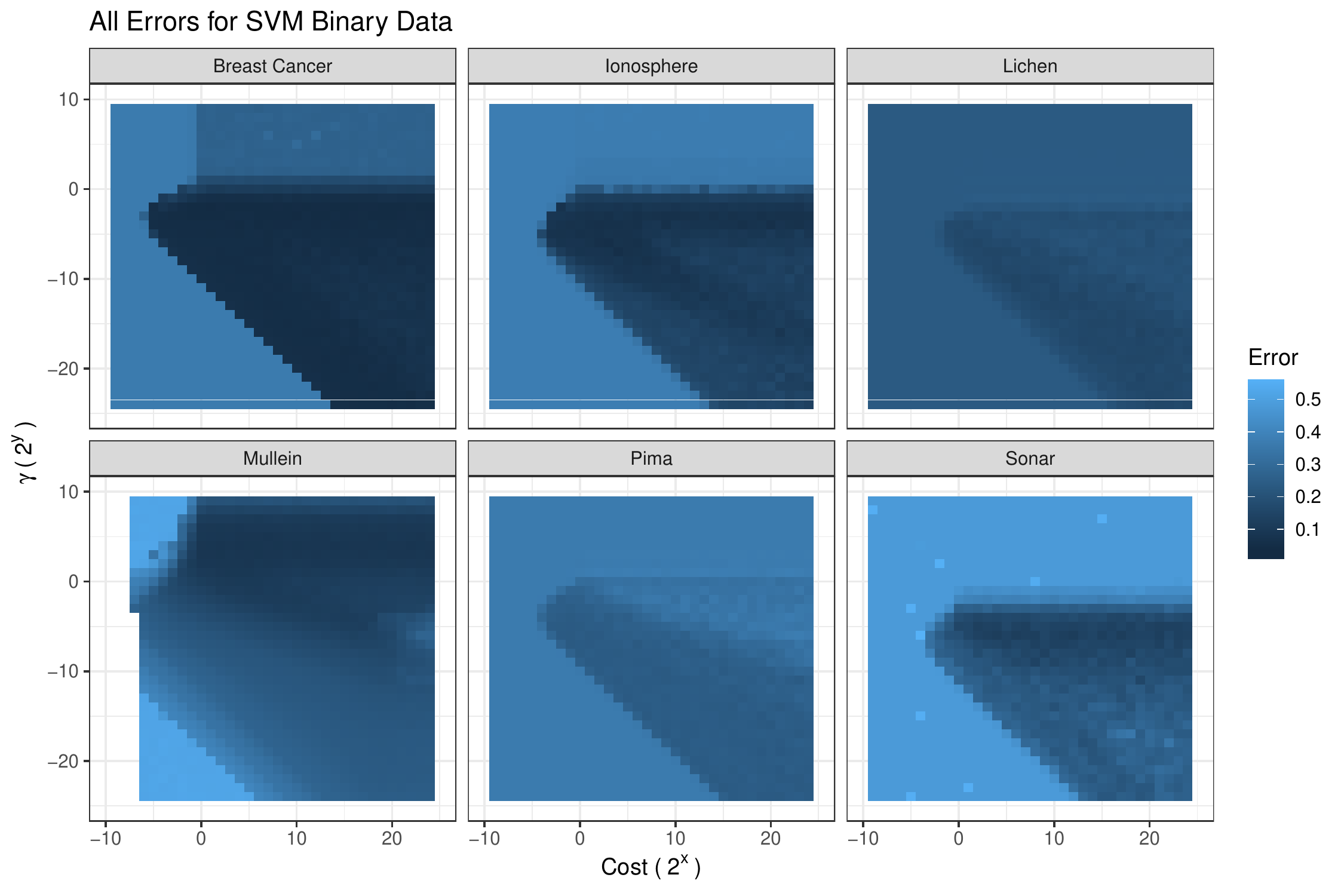}
		\includegraphics[width=0.95\textwidth]{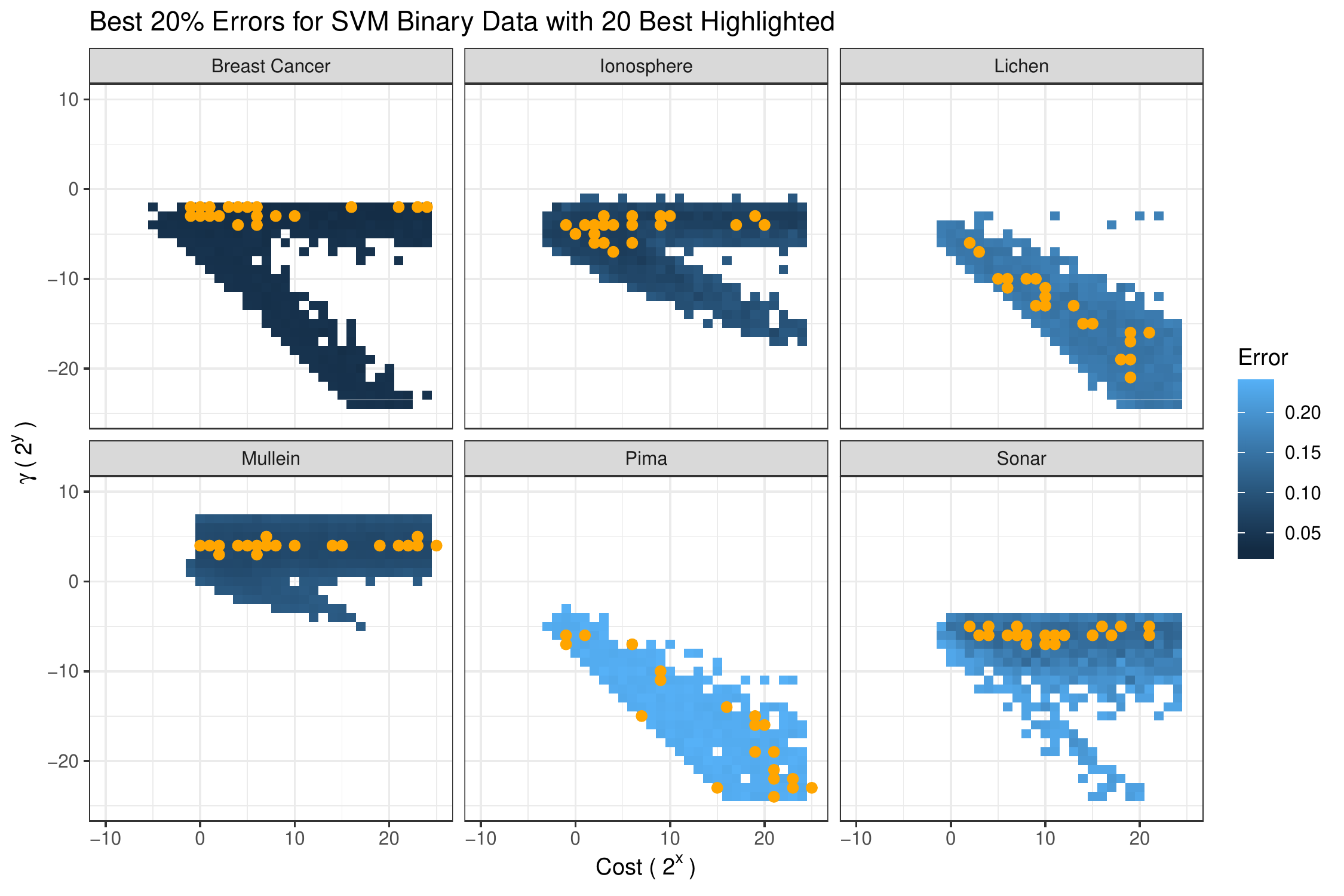}
	\end{center}
		\caption{Error surface plots for support vector machines on datasets with a binary response. Orange dots on the bottom figure represent the best 20 models across the grid.}
		\label{fig:svmerr}
\end{figure}

\begin{figure}[!ht]
	\begin{center}
		\includegraphics[width=\textwidth]{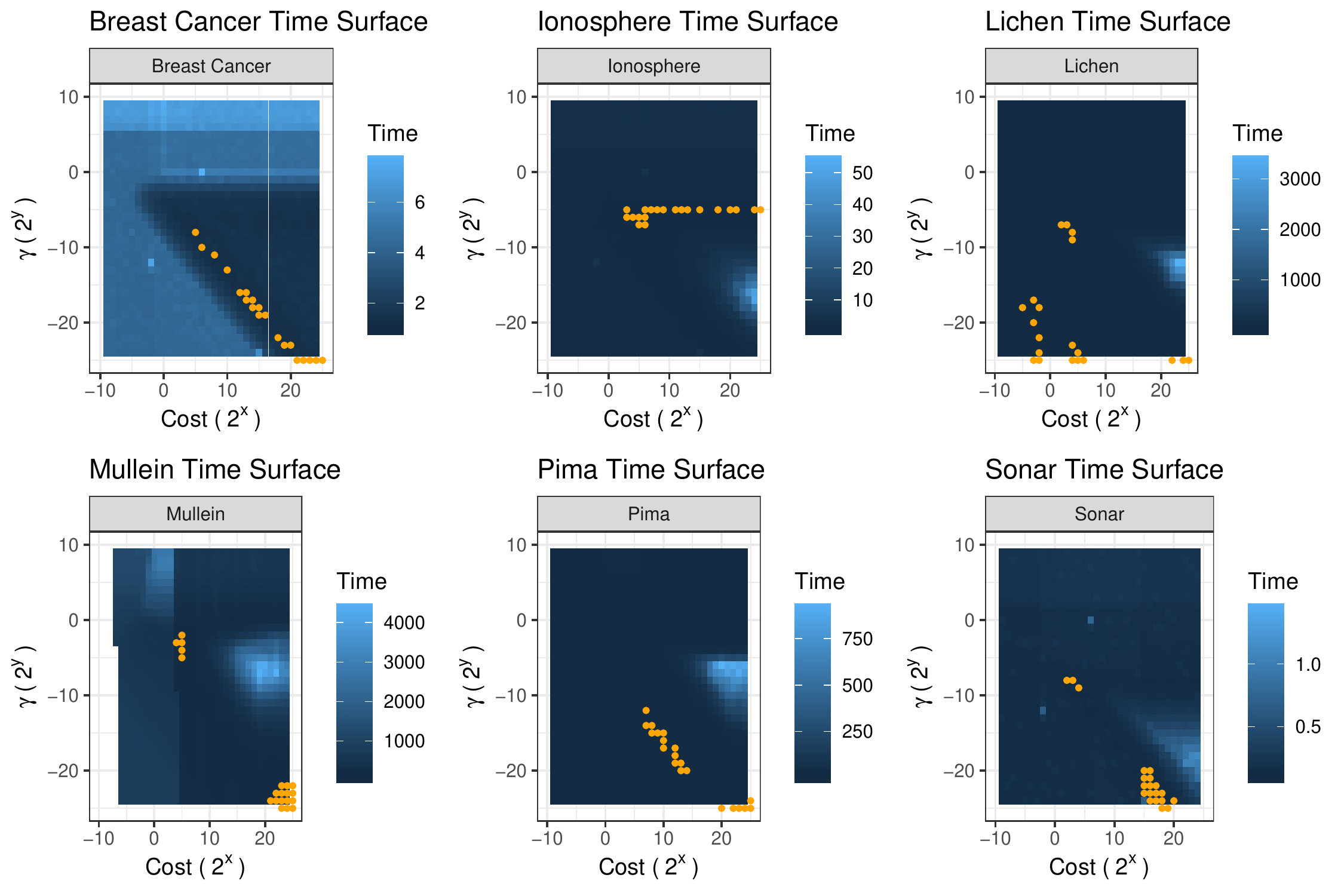}
	\end{center}
		\caption{Computation time surface plots for support vector machines on datasets with a binary response. Orange dots represent the 20 models with the shortest computation times across the grid. Time is in seconds.}
		\label{fig:svmtime}
\end{figure}

Grid searches were done using the data listed in Table \ref{tbl:bindat} over the parameter ranges specified in Tables \ref{tbl:gridrangeB} and \ref{tbl:gridrangeC}. Figure \ref{fig:svmerr} shows the surface of the errors obtained for the classification SVM models. Although a distinct surface emerges across all of the datasets, it is difficult to determine a smaller region where performance is good across all datasets. The grid results were subsetted to include only best 20\% of the errors and to include the best 20 error measures across the entire grid. Figure \ref{fig:svmtime} shows the surface for the computation times across the grid with the fastest 20 computation times highlighted in orange. Surface plots for the other models are included in the supplementary materials.   

The wide distribution of the orange dots in Figures \ref{fig:svmerr} and \ref{fig:svmtime} shows that there are many local minima across the surface. Figure \ref{fig:svmtime} also shows that there are areas in the grid with slow computation times. The best computation times seemed to be in the same grid regions with the best error rates. The MSE and computation time surfaces for the regression models were similar to those for the classification models. Smaller values of $\epsilon$ produced smaller MSEs but also had slower computation times for all datasets. Good error rates with reasonable computation times can be obtained by models with a cost between 1 and 1000 and a $\gamma$ between $2^{-10}$ to $2^{10}$. The best results for regression were seen for values of $\epsilon$ less than 0.5. It is clear from the analysis that cost, $\gamma$, and $\epsilon$ should all be tuned. Although the regression and classification models had similar response surfaces, the tuning parameter spaces selected for data types are slightly different. This is so the subtle differences between each model type can be best utilized. Tables \ref{tbl:gridrangeB} and \ref{tbl:gridrangeC} show the selected tuning parameter spaces for all of the models. Starting values for each of the parameter spaces were also selected from the error surfaces. The starting locations were selected from areas that tend to have low error measures and faster computation times across all datasets. 

The parameter regions for GBM were searched in a similar manner and response surfaces showed patterns in performance. However, none of the error rate or computation time surfaces were smooth, even when examined with multidimensional graphics. Computation times were unilaterally faster with smaller values for all tuning parameters with the exception of shrinkage. The best error measures were found across the range of shrinkage values, so a smaller shrinkage does not always result in a better model fit. Better error rates were also found with fewer than 1000 iterations and often with only about 500 iterations. Good results were seen across the spectrum of tested interaction depths and the different values of the minimum number of observations in the terminal nodes. The areas of best performance varied for each dataset so it was determined that the range of values should be minimally trimmed for those two tuning parameters. As with the SVM analysis, it was clear that it is important to tune all tuning parameters. Tables \ref{tbl:gridrangeB} and \ref{tbl:gridrangeC} show the selected tuning parameter spaces. 

Adaboost was evaluated only for the binary datasets. The best computation times were seen for the smallest number of trees and the smallest tree depths. Shrinkage did not have much impact on computation times. Good error rates were seen across all values of shrinkage that were tested and good models were found for all values of tree depth and the number of iterations. The tuning parameter space was chosen to try to minimize computation time while catching some of the best models for each dataset. Tables \ref{tbl:gridrangeB} and \ref{tbl:gridrangeC} shows the selected tuning parameter spaces. 

Smaller regions than those listed in Tables \ref{tbl:gridrangeB} and \ref{tbl:gridrangeC} were tested during the optimization phase to determine if reducing this region to a smaller area improves computation time with little sacrifice in accuracy. It was found that smaller regions did not decrease computation times for most optimization algorithms and often resulted in larger error measures so the larger parameter space was retained. 


\section{Optimization algorithms}
\label{sec:opt}

Once a parameter space was determined, the parameter spaces for each dataset were searched using 17 different optimization algorithms. Table \ref{tbl:optim} lists the algorithms that were assessed. Ten searches were carried out for each dataset and algorithm. Both the error measure and computation time were computed for each search to determine the stability of the algorithm. If an algorithm was not able to complete 10 runs within a specified time frame for one of the datasets, it was considered a failure for that dataset. 

The R statistical package \cite{R} was used for all optimization computations. Table \ref{tbl:optim} shows the R packages and functions that were used for the optimization. Computation time and error measures were compared and it was assumed that different optimization algorithms may perform better for different model types. A description of each of these algorithms can be found in Table \ref{tbl:optalg} of the supplementary material.

\begin{table}
\caption{Optimization algorithms along with the packages and functions in R that will be used to implement them.}
\begin{center}
    \begin{tabular}{lllll}
    \hline
     Algorithm	&	Type	&	Package	&	Function	\\
    \hline							
     Ant Lion \cite{mirjalili2015ant}	&	Metaheuristic	&	MetaheuristicOpt 	&	ALO \\	
     BOBYQA \cite{powell2009bobyqa}	 	&	Derivative free	&	minqa \cite{minqa}	&	bobyqa	\\
     Dragonfly  	&	Metaheuristic	&	MetaheuristicOpt	&	DA	\\
    \cite{mirjalili2016dragonfly}	&		&		&		\\
     Firefly  \cite{yang2009firefly}	&	Metaheuristic	&	MetaheuristicOpt	&	FFA	\\
     Genetic algorithm\cite{goldberg1999genetic}	&	Metaheuristic	&	GA \cite{GA}	&	ga	\\
     Grasshopper	\cite{saremi2017grasshopper} &	Metaheuristic	&	MetaheuristicOpt	&	GOA	\\
     Grey wolf \cite{mirjalili2014grey}	&	Metaheuristic	&	MetaheuristicOpt	&	GWO	\\
     Hooke-Jeeves \cite{lai2007distributed}	&	Derivative free	&	optimx \cite{optimx}, 	&	hjk,	\\
     	&		&	dfoptim \cite{dfoptim}	&	hjkb	\\
     Improved harmony search	\cite{mahdavi2007improved} &	Metaheuristic	&	MetaheuristicOpt	&	HS	\\
     L-BFGS	\cite{byrd1995limited} &	Quasi-Newton	&	lbfgsb3 \cite{lbfgsb3},	&	lbfgsb3,	\\
     	&		&	stats \cite{R}	&	optim	\\    				
     Moth flame	\cite{mirjalili2015moth} &	Metaheuristic	&	MetaheuristicOpt	&	MFO	\\
     Nelder-Mead \cite{kelley1999iterative}	&	Derivative free	&	dfoptim	&	nmk	\\
     Nonlinear conjugate	gradient \cite{dai2001efficient} &	Gradient 	&	Rcgmin \cite{Rcgmin}	&	Rcgmin	\\
     Particle swarm	\cite{shi1998modified} &	Metaheuristic	&	MetaheuristicOpt	&	PSO	\\
     Sine cosine	\cite{mirjalili2016sca} &	Metaheuristic	&	MetaheuristicOpt	&	SCA	\\
     Spectral projected	\cite{birgin2000nonmonotone} &	Gradient	&	BB 	&	spg	\\
     gradient 	&		&	\cite{BB}	&		\\
     Whale \cite{mirjalili2016whale}	&	Metaheuristic	&	MetaheuristicOpt	&	WOA	\\
    \hline
    \end{tabular}
\end{center}
\label{tbl:optim}
\end{table}

\begin{figure}[!htb]
	\begin{center}
		\includegraphics[width=\textwidth]{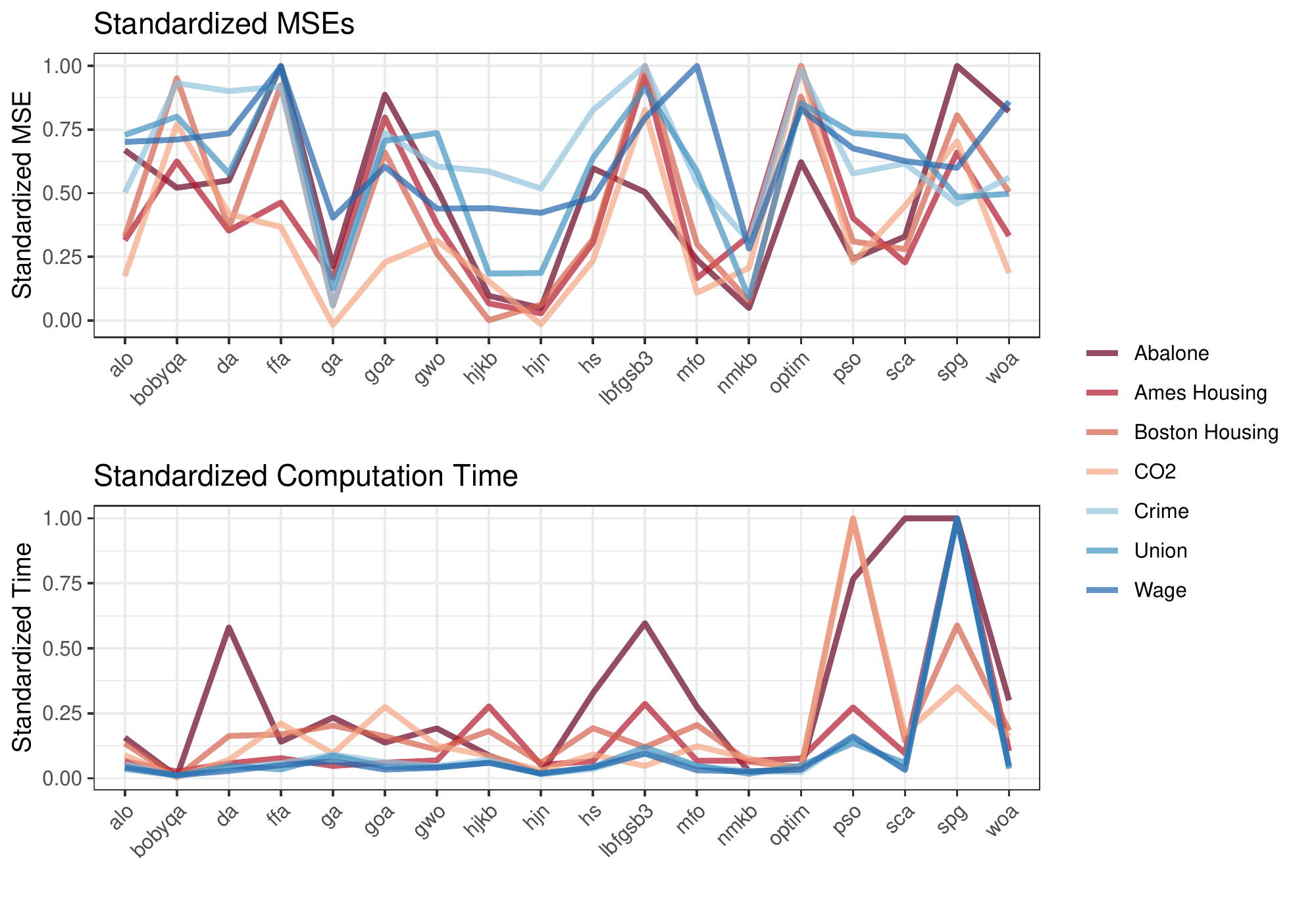}
	\end{center}
		\caption{Standardized optimization results for support vector regression.}
		\label{fig:time-reg-svm}
\end{figure}

\begin{figure}[!htb]
	\begin{center}
		\includegraphics[width=\textwidth]{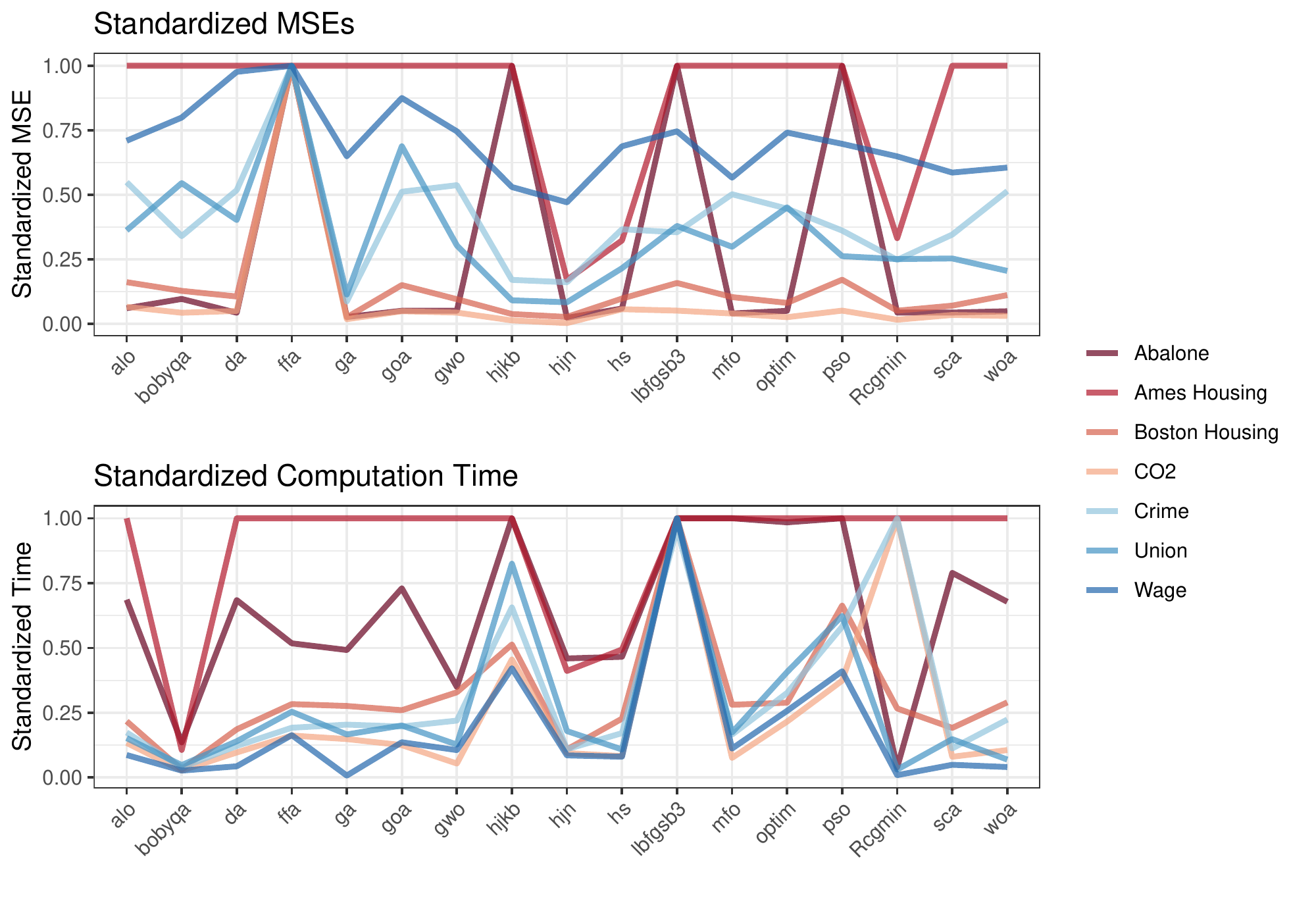}
	\end{center}
		\caption{Standardized optimization results for gradient boosting machine regression.}
		\label{fig:GBM_time-reg-gbm}
\end{figure}

\begin{figure}[!htb]
	\begin{center}
		\includegraphics[width=\textwidth]{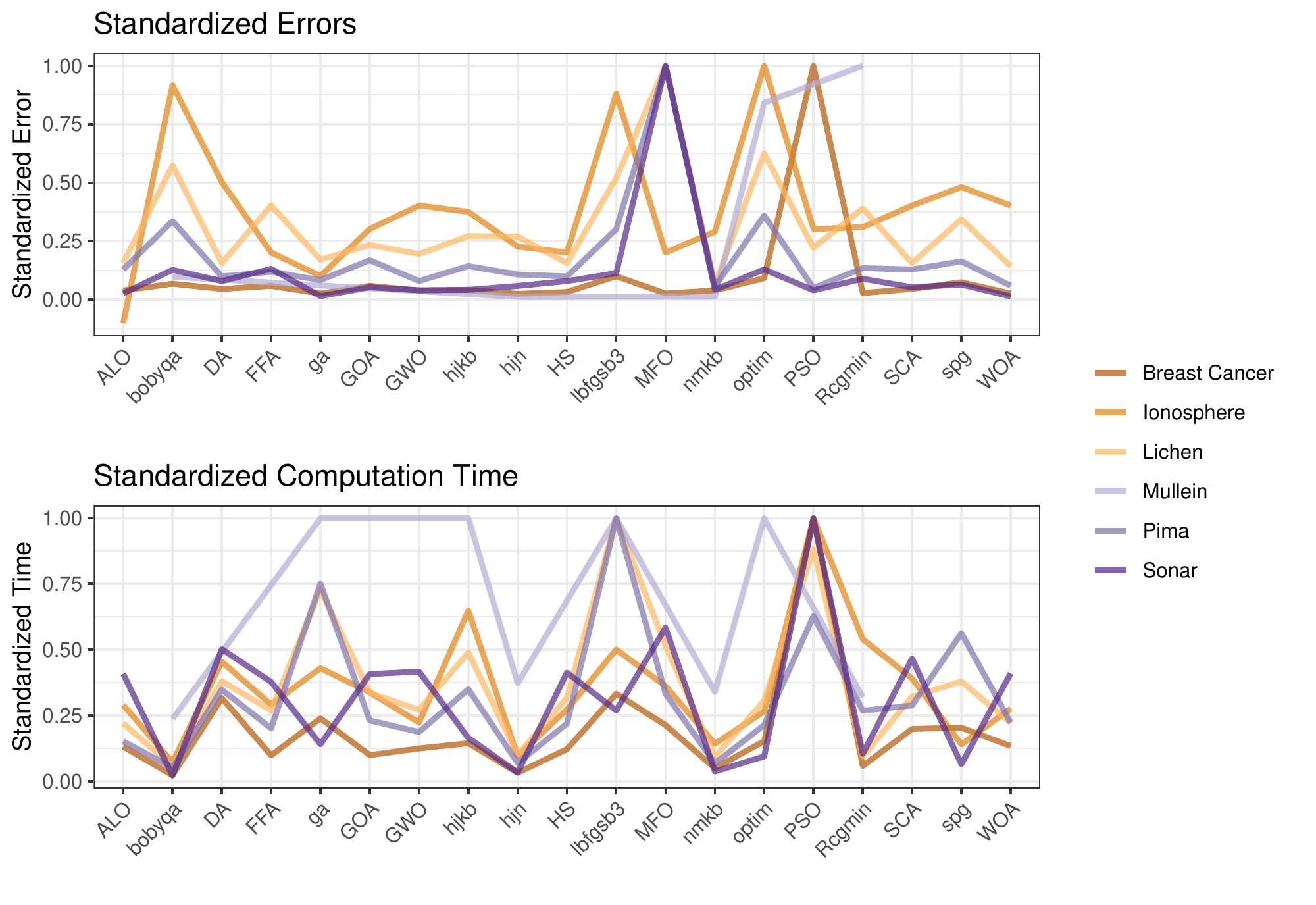}
	\end{center}
		\caption{Standardized optimization results for support vector machines for binary classification.}
		\label{fig:time-bin-svm}
\end{figure}

\begin{figure}[!htb]
	\begin{center}
		\includegraphics[width=\textwidth]{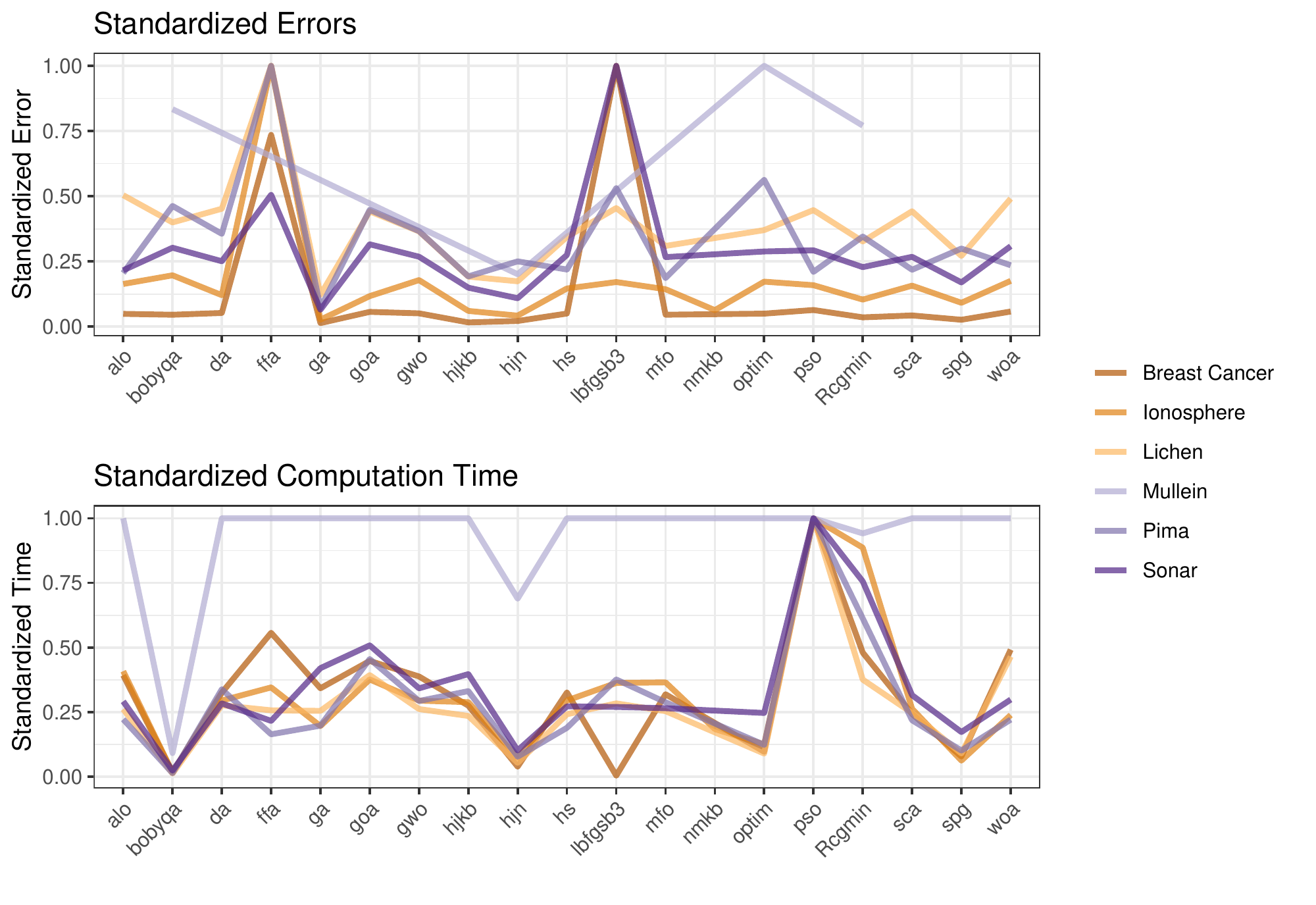}
	\end{center}
		\caption{Standardized optimization results for gradient boosting machines for binary classification.}
		\label{fig:time-bin-gbm}
\end{figure}

\begin{figure}[!htb]
	\begin{center}
		\includegraphics[width=\textwidth]{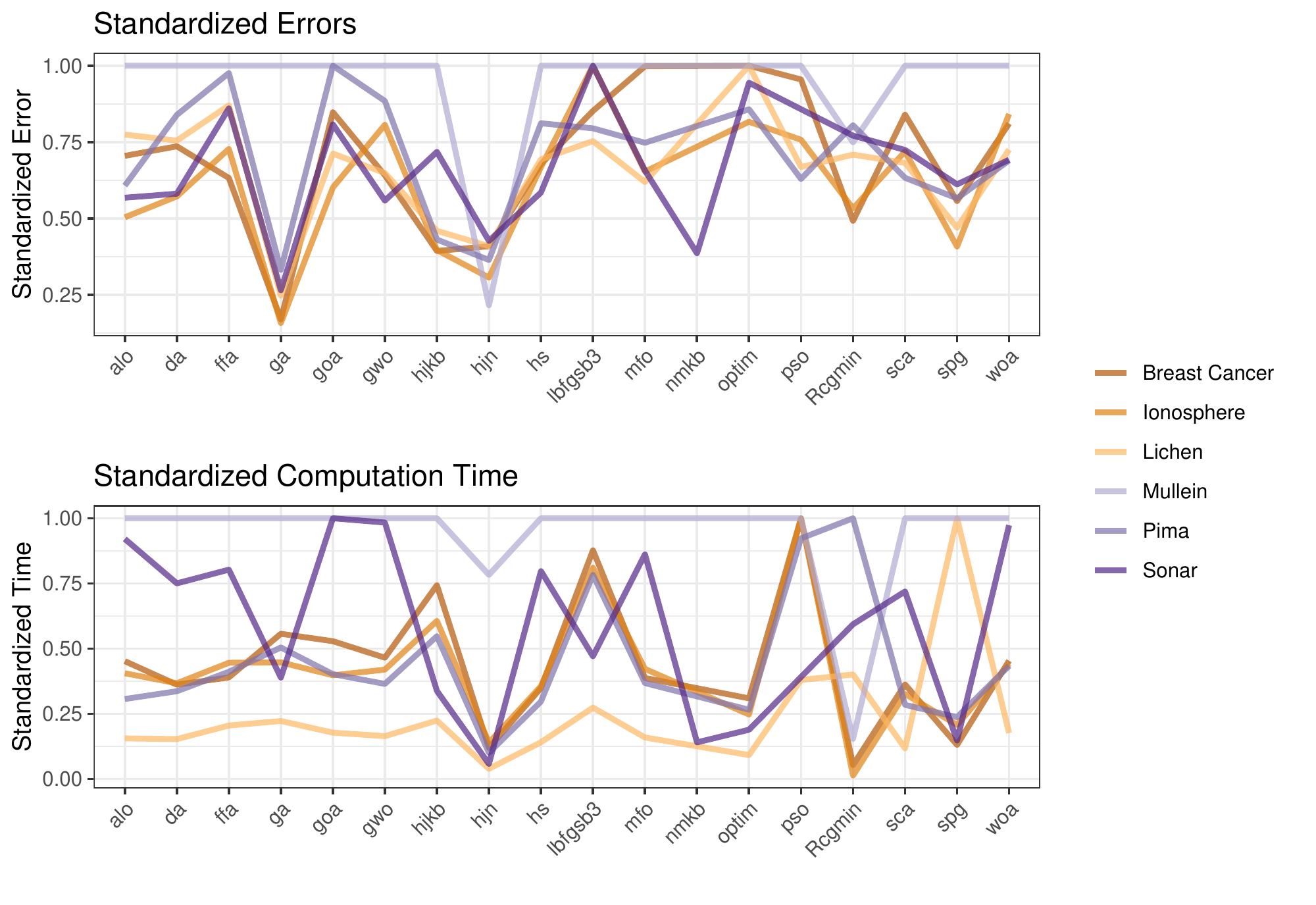}
	\end{center}
		\caption{Standardized optimization results for adaboost models for binary classification.}
		\label{fig:time-bin-ada}
\end{figure}

Table \ref{tbl:optim-res} concisely summarizes the results of all of the optimization algorithms. Figures \ref{fig:time-reg-svm} - \ref{fig:time-bin-ada} show parallel coordinate plots of the results of the optimization tests. Values in the graphs have been standardized by subtracting the minimum measure obtained from the grid search and then dividing by the maximum resulting value. This means that the worst error measure or computation time for each dataset is 1 and the best error measure or computation time is 0. The x-axis lists the R function name to avoid confusion for algorithms that were tested with more than one function. If the computation time was too long to obtain a result, a value of 1 was plotted for both the time and the error measure. 

\begin{table}
\caption{Performance summary of optimization algorithms.}
\begin{center}
    \begin{tabular}{llllll}
        \hline
        Method	&	Error	&	Time	&	Consistency	\\
        \hline							
        \textcolor{blue}{Genetic algorithm}	&	\textcolor{blue}{Good}	&	\textcolor{blue}{Slow}	&	\textcolor{blue}{Consistent}	\\
        Hooke and Jeeves 	&	Varies	&	Varies	&	Inconsistent	\\
        \textcolor{blue}{Hooke and Jeeves B} 	&	\textcolor{blue}{Good}	&	\textcolor{blue}{Good}	&	\textcolor{blue}{Consistent}	\\
        L-BFGS  	&	Good	&	Good	&	Crashes often	\\
        Nocedal-Morales	&	Poor	&	Slow	&	Consistent	\\
        Nonlinear conjugate gradient 	&	Poor	&	Fast	&	Stays at start	\\
        BOBYQA	&	Poor	&	Fast	&	Consistent	\\
        L-BFGS B	&	Poor	&	Fast	&	Consistent	\\
        Spectral projected gradient 	&	Moderate to Poor	&	Good to moderate	&	Inconsistent	\\
        Ant lion	&	Poor	&	Moderate	&	Consistent	\\
        Dragonfly	&	Poor	&	Good	&	Consistent	\\
        Firefly	&	Poor	&	Slow	&	Consistent	\\
        Grasshopper	&	Moderate	&	Moderate	&	Inconsistent	\\
        Grey wolf	&	Poor	&	Moderate	&	Inconsistent	\\
        Harmony search	&	Poor	&	Moderate	&	Inconsistent	\\
        Moth flame	&	Poor	&	Moderate	&	Inconsistent	\\
        Particle swarm	&	Poor	&	Slow	&	Consistent	\\
        Sine cosine	&	Poor	&	Moderate	&	Inconsistent	\\
        Whale optimization	&	Poor	&	Slow	&	Consistent	\\
        \hline
        \end{tabular}
\end{center}
\label{tbl:optim-res}
\end{table}

\section{Discussion}
\label{sec:discuss}

The grid search showed that there are tuning parameter spaces that contain models with small error rates and fast computation times across all of the tested datasets. Areas that have fast computation times for the SVM models include models with good error rates for binary classification. Regression models with SVM showed that this was true for cost and $\gamma$, but not always for $\epsilon$. Tuning parameter spaces for binary and continuous responses were similar, but a smaller region for $\gamma$ can be used when the response is continuous. 

GBM and adaboost have nearly identical tuning parameters, but they behave differently. GBM requires a larger range of trees and interaction depths than adaboost. GBM also requires smaller shrinkage values than adaboost. With both models, a smaller shrinkage was not always better but also did not seem to increase computation time. 

The optimization algorithms listed in Table \ref{tbl:optim} differed markedly in computation time and in their ability to find a set of parameters that produce a good model. We initially thought that a gradient method would perform well for the SVM models because the error surfaces were smooth and that non-gradient based algorithms would be better for GBM and adaboost. However, gradient based methods performed poorly for all models. The non-linear conjugate gradient algorithm had a very fast computation time, but it failed to move from the starting values. This may be an artifact of the \texttt{Rcgmin} function that was used \cite{Rcgmin}, but it is a gradient based function and it is unlikely it will perform well regardless of how it is coded. The Nelder-Meade algorithm had fast computation times and low error rates that rivaled all other algorithms, but it often failed to converge. It is worth exploring this algorithm in another programming language, such as Python, to see if more stable performance can be achieved. The metaheuristic algorithms seem like they would perform well based on the appearance of the error rate surfaces and based on the performance of the genetic algorithm so further investigation in Python or another language may yield better results. The Hooke-Jeeves algorithm consistently produced the best error measures and computation times for all datasets across all model types. The genetic algorithm found the best error rates overall, but computation times were slow. With larger datasets, the computation time of the genetic algorithm was prohibitive. 

The success in finding tuning parameter spaces across multiple datasets and the identification of two optimization algorithms that were able to find well tuned models makes it possible to automate tuning. We wrote an R package, \texttt{EZtune} \cite{Eztune}\cite{dissertation}, that can tune an SVM, GBM, or adaboost model with little input from the user. The user selects a model type in the the main function, \texttt{eztune}, and the function searches the tuning parameter spaces listed in Tables \ref{tbl:gridrangeB} and \ref{tbl:gridrangeC} using either a Hooke-Jeeves or genetic algorithm. The function returns a tuned model, the values for the tuning parameters, and performance metrics. The results of the \texttt{eztune} function were compared to the grid search results and the models that were returned were as good, or nearly as good, as the best model found in the grid search. The benchmarking results are included in Section \ref{sec:s2} of the supplementary material. The \texttt{eztune} function also contains arguments that can substantially decrease computation time with little sacrifice in model performance. The package contains a second function, \texttt{eztune\_cv} that will compute a cross-validation accuracy rate or MSE for any model generated using \texttt{eztune}. 

\section{Conclusions}
\label{sec:conclustions}
This computational work demonstrates that practical tuning parameter spaces can be identified for SVMs and boosted trees that work well across many different datasets. These spaces can increase the speed and efficacy of tuning an SVM or boosted tree. In addition to narrowing the search region, we have demonstrated that the Hooke-Jeeves algorithm and a genetic algorithm can search the tuning parameter space to identify a tuned model that is often as good, or nearly as good, as a model found by a very extensive grid search. The information in this article not only provides insight into tuning these types of models, but it can also be used to create algorithms that automatically tune SVMs and boosted trees well with reasonable computation time. 

\clearpage


\appendix

\renewcommand{\appendixname}{Supplementary Information}

\renewcommand{\thesection}{S.\arabic{section}}    

\renewcommand{\thefigure}{S.\arabic{figure}}
\setcounter{figure}{0}
\renewcommand{\thetable}{S.\arabic{table}}
\setcounter{table}{0}

\begin{center}
{\large\bf SUPPLEMENTARY MATERIAL}
\end{center}
\bigskip
\begin{description}

\item[\ref{sec:s1} Description of optimization algorithms:] This table includes descriptions of each of the optimization algorithms tested in this article. (LaTeX tables)

\item[\ref{sec:s2} Performance tables:] These tables provide additional information about the performance of the different optimization algorithms. (LaTeX tables)

\item[R-package for  EZtune:] R-package EZtune that can implement autotuning of SVMs, GBMs, and adaboost using the Hooke-Jeeves algorithm and genetic algorithm. The package also contains Lichen and Mullein datasets used in the examples in the article. The package is currently available on CRAN and updates are available at https://github.com/jillbo1000/EZtune. (GNU zipped tar file)

\item[Code and data for creating grids and performing optimization tests:] The code and data used to create the error and time response surfaces and the code for testing the optimization algorithms is available at https://github.com/jillbo1000/autotune.

\end{description}

\clearpage

\section{Description of optimization algorithms}
\label{sec:s1}
  \begin{longtable}{llll}
  \caption{List of optimization algorithms used to search tuning parameter spaces with a brief description of each method.} 
  \label{tbl:optalg} \\
        \hline
        Algorithm	&	Type	&	Description	\\
        \hline
        \endfirsthead
        \multicolumn{3}{l}%
        {\tablename\ \thetable\ -- \textit{Continued from previous page}} \\
        \hline
       Algorithm	&	Type	&	Description \\
        \hline
        \endhead
        \hline \multicolumn{3}{r}{\textit{Continued on next page}} \\
        \endfoot
        \hline
        \endlastfoot

    Ant Lion	&	Metaheuristic	&	Based on the hunting mechanisms 	\\
    \cite{mirjalili2015ant}	&		&	of antlions	\\
    \hline
    BOBYQA	&	Derivative free	&	Derivative free optimization by 	\\
    \cite{powell2009bobyqa}	&		&	quadratic approximation	\\
    \hline
    Dragonfly	&	Metaheuristic	&	Based on static and dynamic 	\\
    \cite{mirjalili2016dragonfly}	&		&	swarming behaviors of dragonflies	\\
    \hline
    Firefly	&	Metaheuristic	&	Based on fireflies use of light to 	\\
    \cite{yang2009firefly}	&		&	attract other fireflies	\\
    \hline
    Genetic algorithm	&	Metaheuristic	&	Uses the principles of natural	\\
    \cite{goldberg1999genetic}	&		&	 selection in successive generations 	\\
    	&		&	to find an optimal solution	\\
    \hline
    Grasshopper	&	Metaheuristic	&	Mimics the behavior of 	\\
    \cite{saremi2017grasshopper}	&		&	grasshopper swarms	\\
    \hline
    Grey wolf	&	Metaheuristic	&	Mimics leadership hierarchy 	\\
    \cite{mirjalili2014grey}	&		&	and hunting methods of grey wolves	\\
    \hline
    Hooke-Jeeves	&	Derivative free	&	Pattern search that does a local 	\\
    \cite{lai2007distributed}	&		&	search to find a direction where 	\\
    	&		&	performance improves and then 	\\
    	&		&	moves in that direction making 	\\
    	&		&	larger moves as long as 	\\
    	&		&	improvement continues	\\
    \hline
    Improved harmony search	&	Metaheuristic	&	Mimics the improvisational 	\\
    \cite{mahdavi2007improved}	&		&	process of musicians	\\
    \hline
    L-BFGS	&	Quasi-Newton	&	Second order method that 	\\
    \cite{byrd1995limited}	&		&	estimates the Hessian using 	\\
    	&		&	only recent gradients	\\
    \hline
    Moth flame	&	Metaheuristic	&	Based on the navigation 	\\
    \cite{mirjalili2015moth}	&		&	method of moths called 	\\
    	&		&	transverse orientation	\\
    \hline
    Nelder-Mead 	&	Derivative free	&	Direct search algorithm that 	\\
    \cite{kelley1999iterative}	&		&	generates a simplex from sample	\\
    	&		&	 points, \textit{x}, and uses 	\\
    	&		&	values of \textit{f(x)} at the 	\\
    	&		&	vertices to search for an 	\\
    	&		&	optimal solution	\\
    \hline
    Nonlinear conjugate	&	Gradient 	&	The residual is replaced 	\\
    gradient 	&		&	by a gradient and combined 	\\
    \cite{dai2001efficient}	&		&	with a line search method	\\
    \hline
    Particle swarm	&	Metaheuristic	&	Based on the evolutionary 	\\
    \cite{shi1998modified}	&		&	mechanisms that allows 	\\
    	&		&	organisms to adjust their flying 	\\
    	&		&	based on its own flying 	\\
    	&		&	experience and the experiences	\\
    	&		&	 of its companions	\\
    \hline
    Sine cosine	&	Metaheuristic	&	Creates multiple initial random 	\\
    \cite{mirjalili2016sca}	&		&	possible solutions and requires 	\\
    	&		&	them to fluctuate towards the 	\\
    	&		&	optimal solution using a 	\\
    	&		&	mathematical model based 	\\
    	&		&	on sine and cosine functions	\\
    \hline
    Spectral projected	&	Gradient	&	Uses the spectrum of the 	\\
    gradient 	&		&	underlying Hessian to 	\\
    \cite{birgin2000nonmonotone}	&		&	determine the step lengths 	\\
    	&		&	for gradient descent	\\
    \hline
    Whale 	&	Metaheuristic	&	Mimics the bubble-net 	\\
    \cite{mirjalili2016whale}	&		&	hunting strategy of 	\\
    	&		&	humpback whales	\\
    \hline
      \end{longtable}

\clearpage

\section{Performance tables}
\label{sec:s2}
\begin{sidewaystable}[ph!]

\caption{Average mean squared errors from cross validation model verification and computation times in seconds for support vector regression with EZtune. The best mean squared errors from the grid search are included in the table for reference. Table entries are (cross validated MSE, computation time in seconds).}
\resizebox{\textwidth}{!}{%

    \begin{tabular}{llllllllll}

    \hline
    Optimizer & Type & Abalone & BostonHousing & CO2 & Crime & AmesHousing & Union & Wage \\
    \hline
    Hooke-Jeeves & Resub & (7.386, 918s) & (81.7, 19s) & (72.42, 1s) & (1114, 1s) & (5.42e+09, 346s) & (264.8, 2s) & (2510, 0s) \\
    Hooke-Jeeves & CV = 10 & (4.409, 976s) & (7.941, 156s) & (16.92, 8s) & (659.5, 1s) & (8.61e+08, 1554s) & (76.46, 1s) & (343.8, 1s) \\
    Hooke-Jeeves & CV = 3 & (4.424, 353s) & (8.991, 34s) & (16.87, 2s) & (754.1, 1s) & (8.64e+08, 647s) & (75.41, 1s) & (422.1, 1s) \\
    Hooke-Jeeves & Fast = TRUE & (4.46, 7s) & (9.886, 3s) & (23.5, 1s) & (848.6, 1s) & (9.26e+08, 25s) & (77.69, 1s) & (617.4, 1s) \\
    Hooke-Jeeves & Fast = 0.25 & (4.45, 31s) & (9.797, 2s) & (22.66, 1s) & (1053, 1s) & (9.41e+08, 40s) & (74.05, 1s) & (675.2, 1s) \\
    Hooke-Jeeves & Fast = 0.5 & (4.446, 85s) & (9.482, 5s) & (22.79, 1s) & (827.7, 1s) & (9.51e+08, 79s) & (75.61, 1s) & (526.9, 1s) \\
    Hooke-Jeeves & Fast = 0.75 & (4.459, 132s) & (9.754, 9s) & (21.94, 1s) & (793.8, 1s) & (9.40e+08, 161s) & (76.87, 1s) & (705.9, 1s) \\
    Hooke-Jeeves & Fast = 0.9 & (4.451, 174s) & (9.762, 13s) & (18.79, 1s) & (853.6, 1s) & (9.26e+08, 221s) & (79.69, 1s) & (689.1, 1s) \\
    Hooke-Jeeves & Fast = 100 & (4.462, 6s) & (10.1, 2s) & NA & NA & (9.60e+08, 18s) & NA & NA \\
    Hooke-Jeeves & Fast = 200 & (4.464, 8s) & (9.588, 4s) & NA & NA & (9.64e+08, 25s) & NA & NA \\
    Hooke-Jeeves & Fast = 300 & (4.478, 11s) & (9.194, 9s) & NA & NA & (9.94e+08, 36s) & NA & NA \\
    Hooke-Jeeves & Fast = 400 & (4.454, 14s) & (9.805, 8s) & NA & NA & (9.35e+08, 48s) & NA & NA \\
    \hline
    Genetic Algorithm & Resub & (7.667, 10494s) & (39.06, 27s) & (77.76, 4s) & (1319, 3s) & (4.58e+09, 352s) & (441.4, 3s) & (2396, 2s) \\
    Genetic Algorithm & CV = 10 & (4.459, 34760s) & (7.999, 283s) & (13.32, 26s) & (611, 7s) & (8.57e+08, 1620s) & (73.09, 8s) & (323.9, 3s) \\
    Genetic Algorithm & CV = 3 & (4.514, 8307s) & (8.393, 83s) & (12.82, 8s) & (634.7, 5s) & (9.11e+08, 548s) & (76.99, 4s) & (330.5, 3s) \\
    Genetic Algorithm & Fast = TRUE & (4.62, 40s) & (10.1, 13s) & (14.1, 4s) & (727.5, 4s) & (8.95e+08, 44s) & (83.97, 3s) & (717.1, 3s) \\
    Genetic Algorithm & Fast = 0.25 & (4.574, 252s) & (10.42, 9s) & (13.58, 4s) & (724.9, 4s) & (8.86e+08, 72s) & (116.3, 3s) & (696.8, 3s) \\
    Genetic Algorithm & Fast = 0.5 & (4.586, 973s) & (9.817, 16s) & (12.9, 5s) & (817.4, 4s) & (8.80e+08, 144s) & (77.97, 3s) & (676.3, 3s) \\
    Genetic Algorithm & Fast = 0.75 & (4.581, 2251s) & (9.963, 26s) & (16.38, 5s) & (847, 5s) & (9.54e+08, 194s) & (78.87, 3s) & (784.9, 3s) \\
    Genetic Algorithm & Fast = 0.9 & (4.67, 2659s) & (10.16, 36s) & (18.77, 5s) & (900, 4s) & (9.11e+08, 271s) & (90.47, 3s) & (626.1, 4s) \\
    Genetic Algorithm & Fast = 100 & (4.548, 32s) & (9.378, 8s) & NA & NA & (9.59e+08, 35s) & NA & NA \\
    Genetic Algorithm & Fast = 200 & (4.659, 41s) & (10.11, 11s) & NA & NA & (8.57e+08, 43s) & NA & NA \\
    Genetic Algorithm & Fast = 300 & (4.581, 53s) & (9.699, 21s) & NA & NA & (9.15e+08, 53s) & NA & NA \\
    Genetic Algorithm & Fast = 400 & (4.566, 71s) & (8.84, 29s) & NA & NA & (8.80e+08, 77s) & NA & NA \\
    \hline
    Best Grid &  & 4.389 & 7.183 & 10.57 & 487.2 & 7.37e+08 & 62.7 & 4.71e-03 \\
    \hline
    \end{tabular}}
\label{tbl:svr-results}
\end{sidewaystable}

\begin{sidewaystable}[ph!]

\caption{Average mean squared errors from cross validation model verification and computation times in seconds for gradient boosting regression with EZtune. The best mean squared errors from the grid search are included in the table for reference.  Table entries are (cross validated MSE, computation time in seconds).}
\resizebox{\textwidth}{!}{%

    \begin{tabular}{llllllllll}

    \hline
    Optimizer & Type & Abalone & BostonHousing & CO2 & Crime & AmesHousing & Union & Wage \\
    \hline
    Hooke-Jeeves & Resub & (5.659, 1832s) & (8.036, 468s) & (8.409, 13s) & (570.7, 13s) & (7.63e+08, 4154s) & (131.4, 5s) & (2201, 4s) \\
    Hooke-Jeeves & CV = 10 & (4.588, 11366s) & (7.821, 2657s) & (5.281, 103s) & (580.2, 102s) & (7.15e+08, 22685s) & (94.73, 37s) & (1455, 39s) \\
    Hooke-Jeeves & CV = 3 & (4.579, 2515s) & (7.664, 668s) & (6.567, 25s) & (550.4, 20s) & (6.92e+08, 5539s) & (96.38, 8s) & NA \\
    Hooke-Jeeves & Fast = TRUE & (4.961, 163s) & (8.47, 144s) & (7.174, 6s) & (572.8, 5s) & (7.18e+08, 527s) & (94.61, 2s) & NA \\
    Hooke-Jeeves & Fast = 0.25 & (4.721, 423s) & (7.889, 90s) & (5.505, 1s) & (569.8, 1s) & (6.67e+08, 794s) & (107.1, 0s) & NA \\
    Hooke-Jeeves & Fast = 0.5 & (4.573, 706s) & (7.878, 167s) & (7.017, 6s) & (645.6, 5s) & (6.74e+08, 1577s) & (98.26, 2s) & NA \\
    Hooke-Jeeves & Fast = 0.75 & (4.583, 881s) & (8.206, 253s) & (5.355, 10s) & (628.4, 8s) & (7.09e+08, 1907s) & (105.4, 3s) & NA \\
    Hooke-Jeeves & Fast = 0.9 & (4.657, 1193s) & (8.051, 322s) & (6.248, 12s) & (607.4, 11s) & (7.06e+08, 2534s) & (99.76, 4s) & (1744, 3s) \\
    Hooke-Jeeves & Fast = 100 & (4.928, 104s) & (7.923, 63s) & NA & NA & (7.03e+08, 261s) & NA & NA \\
    Hooke-Jeeves & Fast = 200 & (4.947, 161s) & (8.131, 144s) & NA & NA & (7.00e+08, 519s) & NA & NA \\
    Hooke-Jeeves & Fast = 300 & (5.015, 182s) & (7.728, 209s) & NA & NA & (7.26e+08, 736s) & NA & NA \\
    Hooke-Jeeves & Fast = 400 & (4.97, 199s) & (8.176, 253s) & NA & NA & (7.06e+08, 981s) & NA & NA \\
    \hline
    Genetic Algorithm & Resub & (5.778, 4067s) & (8.082, 611s) & (9.286, 34s) & (594.7, 20s) & (8.03e+08, 5941s) & (133.2, 15s) & NA \\
    Genetic Algorithm & CV = 10 & (4.571, 12460s) & (7.987, 5843s) & (6.343, 179s) & (570.2, 194s) & (7.00e+08, 58628s) & (99.73, 48s) & (1751, 36s) \\
    Genetic Algorithm & CV = 3 & (4.581, 3549s) & (7.778, 1398s) & (6.517, 53s) & (556, 42s) & (6.58e+08, 12907s) & (111.3, 14s) & NA \\
    Genetic Algorithm & Fast = TRUE & (4.726, 209s) & (7.965, 259s) & (6.882, 13s) & (580.7, 10s) & (7.19e+08, 927s) & (117.5, 6s) & NA \\
    Genetic Algorithm & Fast = 0.25 & (4.588, 567s) & (8.097, 144s) & (7.975, 2s) & NA & (6.72e+08, 1649s) & (124.9, 2s) & NA \\
    Genetic Algorithm & Fast = 0.5 & (4.598, 1098s) & (7.808, 419s) & (7.059, 15s) & (544.5, 12s) & (7.26e+08, 2780s) & (121.6, 4s) & NA \\
    Genetic Algorithm & Fast = 0.75 & (4.641, 1489s) & (7.853, 447s) & (6.758, 20s) & (573.4, 18s) & (7.36e+08, 5229s) & (126.1, 9s) & NA \\
    Genetic Algorithm & Fast = 0.9 & (4.688, 1814s) & (8.379, 538s) & (7.084, 25s) & (580.7, 22s) & (7.40e+08, 4433s) & (119.2, 10s) & (2118, 7s) \\
    Genetic Algorithm & Fast = 100 & (4.656, 169s) & (8.207, 113s) & NA & NA & (7.10e+08, 469s) & NA & NA \\
    Genetic Algorithm & Fast = 200 & (4.69, 221s) & (7.968, 326s) & NA & NA & (6.98e+08, 1052s) & NA & NA \\
    Genetic Algorithm & Fast = 300 & (4.738, 273s) & (8.039, 484s) & NA & NA & (7.24e+08, 1524s) & NA & NA \\
    Genetic Algorithm & Fast = 400 & (4.681, 303s) & (7.834, 603s) & NA & NA & (6.86e+08, 1580s) & NA & NA \\
    \hline
    Best Grid &  & 4.442 & 6.8 & 4.568 & 391.7 & 6.04e+08 & 82.86 & 0.0262 \\
    \hline
    \end{tabular}}
\label{tbl:gbr-results}
\end{sidewaystable}

\begin{sidewaystable}[ph!]

\caption{Average classification errors from cross validation model verification and computation times in seconds for support vector classification with EZtune. The best classification errors from the grid search are included in the table for reference. Table entries are (cross validated error rate, computation time in seconds).}
\resizebox{\textwidth}{!}{%

    \begin{tabular}{llllllllll}

    \hline
    Optimizer & Type & BreastCancer & Ionosphere & Lichen & Mullein & Pima & Sonar \\
    \hline
    Hooke-Jeeves & Resub & (0.0455, 2s) & (0.0521, 2s) & (0.1821, 11s) & (0.0581, 6998s) & (0.3031, 7s) & (0.1327, 12s) \\
    Hooke-Jeeves & CV = 10 & (0.0312, 5s) & (0.0544, 7s) & (0.1446, 44s) & (0.0574, 70130s) & (0.2363, 26s) & (0.1197, 48s) \\
    Hooke-Jeeves & CV = 3 & (0.0338, 2s) & (0.0587, 3s) & (0.1538, 20s) & (0.0575, 18827s) & (0.2384, 10s) & (0.1274, 26s) \\
    Hooke-Jeeves & Fast = TRUE & (0.0372, 1s) & (0.0587, 2s) & (0.1599, 4s) & (0.1748, 70s) & (0.2384, 1s) & (0.1264, 15s) \\
    Hooke-Jeeves & Fast = 0.25 & (0.0379, 1s) & (0.0581, 2s) & (0.1693, 3s) & (0.0783, 539s) & (0.2393, 1s) & (0.1274, 13s) \\
    Hooke-Jeeves & Fast = 0.5 & (0.0401, 1s) & (0.057, 2s) & (0.164, 5s) & (0.0581, 2745s) & (0.2336, 3s) & (0.1264, 15s) \\
    Hooke-Jeeves & Fast = 0.75 & (0.0379, 1s) & (0.055, 3s) & (0.1705, 8s) & (0.0576, 5128s) & (0.2534, 4s) & (0.1298, 17s) \\
    Hooke-Jeeves & Fast = 0.9 & (0.0397, 1s) & (0.0544, 2s) & (0.1681, 11s) & (0.0681, 7179s) & (0.2436, 5s) & (0.1308, 19s) \\
    Hooke-Jeeves & Fast = 100 & (0.0391, 1s) & (0.0524, 2s) & (0.1623, 3s) & (0.1645, 57s) & (0.2374, 1s) & (0.125, 15s) \\
    Hooke-Jeeves & Fast = 200 & (0.0351, 1s) & (0.0553, 2s) & (0.174, 4s) & (0.169, 64s) & (0.243, 1s) & (0.1298, 16s) \\
    Hooke-Jeeves & Fast = 300 & (0.0365, 1s) & (0.0524, 2s) & (0.16, 5s) & (0.1627, 70s) & (0.2417, 2s) & NA \\
    Hooke-Jeeves & Fast = 400 & (0.0344, 1s) & NA & (0.1582, 6s) & (0.1563, 80s) & (0.2435, 2s) & NA \\
    \hline
    Genetic Algorithm & Resub & (0.053, 8s) & (0.1598, 10s) & (0.2098, 45s) & (0.0578, 13064s) & (0.3234, 25s) & (0.3111, 58s) \\
    Genetic Algorithm & CV = 10 & (0.0321, 26s) & (0.0481, 28s) & (0.1496, 227s) & (0.0583, 175904s) & (0.2302, 248s) & (0.1212, 176s) \\
    Genetic Algorithm & CV = 3 & (0.0327, 11s) & (0.055, 15s) & (0.1494, 84s) & (0.0587, 58531s) & (0.2355, 72s) & (0.1212, 105s) \\
    Genetic Algorithm & Fast = TRUE & (0.0347, 5s) & (0.0575, 9s) & (0.1601, 14s) & (0.1825, 137s) & (0.2322, 8s) & (0.1269, 80s) \\
    Genetic Algorithm & Fast = 0.25 & (0.0362, 4s) & (0.0584, 8s) & (0.1526, 15s) & (0.075, 1338s) & (0.2359, 6s) & (0.1298, 67s) \\
    Genetic Algorithm & Fast = 0.5 & (0.0329, 5s) & (0.0581, 9s) & (0.1487, 24s) & (0.0608, 5163s) & (0.2371, 15s) & (0.126, 76s) \\
    Genetic Algorithm & Fast = 0.75 & (0.0378, 6s) & (0.0601, 9s) & (0.1606, 38s) & (0.0587, 13604s) & (0.2384, 25s) & (0.1173, 76s) \\
    Genetic Algorithm & Fast = 0.9 & (0.0394, 6s) & (0.0655, 8s) & (0.1568, 41s) & (0.059, 13408s) & (0.244, 38s) & (0.1212, 75s) \\
    Genetic Algorithm & Fast = 100 & (0.0397, 4s) & (0.0604, 8s) & (0.1662, 11s) & (0.1898, 117s) & (0.2348, 5s) & (0.1269, 74s) \\
    Genetic Algorithm & Fast = 200 & (0.0354, 5s) & (0.0621, 10s) & (0.1539, 15s) & (0.1813, 126s) & (0.2357, 8s) & (0.1231, 64s) \\
    Genetic Algorithm & Fast = 300 & (0.0359, 4s) & (0.0644, 9s) & (0.1551, 17s) & (0.1786, 142s) & (0.2353, 12s) & NA \\
    Genetic Algorithm & Fast = 400 & (0.0329, 6s) & NA & (0.1515, 22s) & (0.1709, 202s) & (0.232, 15s) & NA \\
\hline
    Best Grid &  & 0.0234 & 0.0427 & 0.131 & 0.0682 & 0.2174 & 0.101 \\ \hline
    \end{tabular}}
\label{tbl:svm-results}
\end{sidewaystable}

\begin{sidewaystable}[ph!]

\caption{Average classification errors from cross validation model verification and computation times in seconds for gradient boosting classification with EZtune. The best classification errors from the grid search are included in the table for reference. Table entries are (cross validated error rate, computation time in seconds).}
\resizebox{\textwidth}{!}{%

    \begin{tabular}{llllllllll}

    \hline
    Optimizer & Type & BreastCancer & Ionosphere & Lichen & Mullein & Pima & Sonar \\
    \hline
    Hooke-Jeeves & Resub & (0.0303, 30s) & (0.0681, 58s) & (0.1606, 124s) & (0.0779, 6388s) & (0.269, 47s) & (0.1341, 347s) \\
    Hooke-Jeeves & CV = 10 & (0.0313, 331s) & (0.0698, 667s) & (0.163, 1529s) & (0.0787, 54699s) & (0.2587, 372s) & (0.1279, 4171s) \\
    Hooke-Jeeves & CV = 3 & (0.0318, 79s) & (0.067, 157s) & (0.1629, 460s) & (0.0786, 11786s) & (0.2577, 107s) & (0.1308, 1081s) \\
    Hooke-Jeeves & Fast = TRUE & (0.0319, 15s) & (0.0712, 47s) & (0.1592, 48s) & (0.1399, 115s) & (0.2665, 17s) & (0.1457, 263s) \\
    Hooke-Jeeves & Fast = 0.25 & (0.0297, 14s) & (0.0687, 23s) & (0.1617, 54s) & (0.0939, 1070s) & (0.2642, 15s) & (0.1428, 107s) \\
    Hooke-Jeeves & Fast = 0.5 & (0.0293, 21s) & (0.0675, 42s) & (0.1581, 114s) & (0.089, 2589s) & (0.2585, 23s) & (0.1288, 275s) \\
    Hooke-Jeeves & Fast = 0.75 & (0.0331, 27s) & (0.0718, 68s) & (0.163, 145s) & (0.0836, 4395s) & (0.2643, 34s) & (0.1428, 381s) \\
    Hooke-Jeeves & Fast = 0.9 & (0.0319, 27s) & (0.0672, 77s) & (0.1611, 152s) & (0.101, 3720s) & (0.2646, 47s) & (0.126, 420s) \\
    Hooke-Jeeves & Fast = 100 & (0.0309, 9s) & (0.0675, 27s) & (0.1594, 30s) & (0.1207, 99s) & (0.2651, 11s) & (0.1327, 247s) \\
    Hooke-Jeeves & Fast = 200 & (0.0312, 15s) & (0.0704, 46s) & (0.1623, 50s) & (0.1217, 137s) & (0.2604, 16s) & (0.1428, 401s) \\
    Hooke-Jeeves & Fast = 300 & (0.0318, 18s) & (0.0655, 65s) & (0.1618, 82s) & (0.1131, 166s) & (0.2613, 22s) & NA \\
    Hooke-Jeeves & Fast = 400 & (0.0306, 27s) & NA & (0.164, 106s) & (0.1062, 187s) & (0.2642, 28s) & NA \\
    \hline
    Genetic Algorithm & Resub & (0.0328, 224s) & (0.0724, 438s) & (0.1585, 1100s) & (0.0703, 17910s) & (0.2747, 266s) & (0.1524, 466s) \\
    Genetic Algorithm & CV = 10 & (0.0294, 2192s) & (0.0644, 4026s) & (0.154, 11211s) & (0.0705, 159183s) & (0.2423, 1609s) & (0.1337, 4398s) \\
    Genetic Algorithm & CV = 3 & (0.0324, 609s) & (0.0692, 820s) & (0.1574, 2654s) & (0.0712, 62064s) & (0.2401, 458s) & (0.1361, 809s) \\
    Genetic Algorithm & Fast = TRUE & (0.0312, 104s) & (0.0695, 241s) & (0.1549, 231s) & (0.1113, 784s) & (0.2495, 80s) & (0.1394, 1152s) \\
    Genetic Algorithm & Fast = 0.25 & (0.0322, 84s) & (0.0675, 115s) & (0.156, 279s) & (0.0746, 6952s) & (0.243, 71s) & (0.1577, 373s) \\
    Genetic Algorithm & Fast = 0.5 & (0.0312, 149s) & (0.0692, 221s) & (0.1575, 712s) & (0.071, 14104s) & (0.2448, 129s) & (0.1457, 941s) \\
    Genetic Algorithm & Fast = 0.75 & (0.0322, 196s) & (0.0738, 379s) & (0.1625, 985s) & (0.0729, 36854s) & (0.2458, 190s) & (0.1481, 1900s) \\
    Genetic Algorithm & Fast = 0.9 & (0.0316, 215s) & (0.0681, 364s) & (0.1585, 1050s) & (0.0709, 34875s) & (0.2544, 222s) & (0.1495, 1875s) \\
    Genetic Algorithm & Fast = 100 & (0.0299, 58s) & (0.0652, 128s) & (0.1565, 142s) & (0.0957, 462s) & (0.2454, 41s) & (0.1375, 1155s) \\
    Genetic Algorithm & Fast = 200 & (0.0309, 103s) & (0.0709, 291s) & (0.159, 344s) & (0.1126, 735s) & (0.2402, 72s) & (0.137, 2034s) \\
    Genetic Algorithm & Fast = 300 & (0.031, 159s) & (0.0687, 387s) & (0.1558, 529s) & (0.089, 1010s) & (0.2406, 95s) & NA \\
    Genetic Algorithm & Fast = 400 & (0.0313, 210s) & NA & (0.1558, 674s) & (0.1026, 1158s) & (0.2443, 112s) & NA \\
    \hline
    Best Grid &  & 0.022 & 0.0484 & 0.1333 & 0.0733 & 0.2214 & 0.0962 \\
    \hline
    \end{tabular}}
\label{tbl:gbm-results}
\end{sidewaystable}

\begin{sidewaystable}[ph!]

\caption{Average classification errors from cross validation model verification and computation times in seconds for adaboost with EZtune. The best classification errors from the grid search are included in the table for reference. Table entries are (cross validated error rate, computation time in seconds).}
\resizebox{\textwidth}{!}{%

    \begin{tabular}{llllllllll}

    \hline
    Optimizer & Type & BreastCancer & Ionosphere & Lichen & Mullein & Pima & Sonar \\
    \hline
    Hooke-Jeeves & Resub & (0.0347, 131s) & (0.0738, 207s) & (0.1689, 381s) & (0.1254, 12601s) & (0.2786, 1216s) & (0.1668, 234s) \\
    Hooke-Jeeves & CV = 10 & (0.0351, 1568s) & (0.0812, 2145s) & (0.1615, 3472s) & (0.1265, 123910s) & (0.2452, 8702s) & (0.1659, 3099s) \\
    Hooke-Jeeves & CV = 3 & (0.0344, 455s) & (0.0781, 662s) & (0.1615, 1086s) & (0.1255, 27970s) & (0.2409, 2596s) & (0.1553, 851s) \\
    Hooke-Jeeves & Fast = TRUE & (0.0357, 114s) & (0.0755, 221s) & (0.165, 299s) & (0.1795, 528s) & (0.2518, 709s) & (0.1663, 322s) \\
    Hooke-Jeeves & Fast = 0.25 & (0.034, 114s) & (0.0789, 210s) & (0.1708, 298s) & (0.1298, 2964s) & (0.249, 672s) & (0.1582, 265s) \\
    Hooke-Jeeves & Fast = 0.5 & (0.036, 148s) & (0.0764, 198s) & (0.1638, 322s) & (0.1272, 6286s) & (0.2574, 778s) & (0.1524, 315s) \\
    Hooke-Jeeves & Fast = 0.75 & (0.0334, 163s) & (0.0789, 218s) & (0.1599, 386s) & (0.1292, 7831s) & (0.2479, 909s) & (0.1562, 302s) \\
    Hooke-Jeeves & Fast = 0.9 & (0.0354, 145s) & (0.0772, 245s) & (0.1658, 370s) & (0.1327, 9046s) & (0.2496, 940s) & (0.1471, 312s) \\
    Hooke-Jeeves & Fast = 100 & (0.0366, 108s) & (0.0778, 174s) & (0.1667, 227s) & (0.1874, 412s) & (0.2543, 595s) & (0.1591, 316s) \\
    Hooke-Jeeves & Fast = 200 & (0.034, 124s) & (0.0766, 239s) & (0.1682, 220s) & (0.1864, 445s) & (0.2491, 767s) & (0.1639, 228s) \\
    Hooke-Jeeves & Fast = 300 & (0.0351, 128s) & (0.0795, 232s) & (0.1667, 310s) & (0.1942, 427s) & (0.2418, 705s) & NA \\
    Hooke-Jeeves & Fast = 400 & (0.0348, 150s) & NA & (0.1687, 297s) & (0.1802, 523s) & (0.247, 703s) & NA \\
    \hline
    Genetic Algorithm & Resub & (0.0306, 523s) & (0.0718, 847s) & (0.1582, 1999s) & (0.1137, 51756s) & (0.276, 4191s) & (0.1404, 1070s) \\
    Genetic Algorithm & CV = 10 & (0.0327, 5165s) & (0.0704, 12083s) & (0.1635, 22474s) & (0.1282, 224020s) & NA & (0.1341, 14681s) \\
    Genetic Algorithm & CV = 3 & (0.0318, 1772s) & (0.0687, 2917s) & (0.1573, 5970s) & (0.1202, 77899s) & (0.2607, 12742s) & (0.1365, 4128s) \\
    Genetic Algorithm & Fast = TRUE & (0.031, 568s) & (0.0724, 793s) & (0.1619, 1228s) & (0.1412, 1837s) & (0.2698, 3064s) & (0.1351, 1358s) \\
    Genetic Algorithm & Fast = 0.25 & (0.0328, 528s) & (0.0698, 738s) & (0.1587, 1305s) & (0.1265, 11429s) & (0.2642, 3588s) & (0.1413, 1120s) \\
    Genetic Algorithm & Fast = 0.5 & (0.0321, 641s) & (0.0724, 848s) & (0.1543, 1395s) & (0.1281, 21992s) & (0.2618, 3508s) & (0.1457, 1210s) \\
    Genetic Algorithm & Fast = 0.75 & (0.03, 676s) & (0.0709, 1070s) & (0.1611, 1770s) & (0.125, 35719s) & (0.2674, 4699s) & (0.1317, 1332s) \\
    Genetic Algorithm & Fast = 0.9 & (0.0316, 473s) & (0.0709, 794s) & (0.1571, 2247s) & (0.1277, 47479s) & (0.2661, 5161s) & (0.1308, 1283s) \\
    Genetic Algorithm & Fast = 100 & (0.0335, 428s) & (0.0729, 944s) & (0.1608, 933s) & (0.1383, 1337s) & (0.2745, 2580s) & (0.1418, 1099s) \\
    Genetic Algorithm & Fast = 200 & (0.0309, 466s) & (0.0721, 825s) & (0.157, 1165s) & (0.1359, 1729s) & (0.2682, 3134s) & (0.1413, 1137s) \\
    Genetic Algorithm & Fast = 300 & (0.0313, 821s) & (0.0746, 859s) & (0.1568, 1502s) & (0.1271, 2433s) & (0.2755, 3895s) & NA \\
    Genetic Algorithm & Fast = 400 & (0.0324, 807s) & NA & (0.1599, 1576s) & (0.1294, 2619s) & (0.2669, 4392s) & NA \\
    \hline
    Best Grid &  & 0.019 & 0.0484 & 0.125 & 0.0814 & 0.2109 & 0.0865 \\
    \hline
    \end{tabular}}
\label{tbl:ada-results}
\end{sidewaystable}

\clearpage


\bibliographystyle{unsrt}  
\bibliography{tuning}

\end{document}